\documentclass[letterpaper, 10 pt, conference]{ieeeconf}  

\makeatletter
\let\NAT@parse\undefined
\makeatother

\IEEEoverridecommandlockouts                              

\usepackage{amsmath} 

\usepackage{graphicx}
\usepackage{caption}
\usepackage{subcaption}
\usepackage{tabularx}
\usepackage{bm}
\usepackage{amsfonts}
\usepackage{hyperref}
\usepackage{color}

\newsavebox{\measurebox}

\DeclareMathOperator*{\argmax}{arg\!\max}
\DeclareMathOperator{\diag}{diag}

\title{\LARGE \bf
Sparse Gaussian Processes for Continuous-Time Trajectory Estimation on Matrix Lie Groups
}

\author{Jing Dong, Byron Boots, and Frank Dellaert
\thanks{J. Dong, B. Boots, and F. Dellaert are with the
College of Computing, Georgia Institute of Technology, USA.
\texttt{jdong@gatech.edu, \{bboots,frank\}@cc.gatech.edu}}
}

\begin{document}

\maketitle
\thispagestyle{empty}
\pagestyle{empty}

\begin{abstract}
Continuous-time trajectory representations are a powerful tool 
that can be used to address several issues in many practical 
simultaneous localization and mapping (SLAM) scenarios,  
like continuously collected measurements distorted by robot motion, or 
during with asynchronous sensor measurements. 
Sparse Gaussian processes (GP) allow for a probabilistic non-parametric 
trajectory representation that enables fast trajectory estimation by sparse GP regression. 
However, previous approaches are limited to dealing with vector space
representations of state only.
In this technical report we extend the work by Barfoot \textit{et al.}~\cite{Barfoot14rss} 
to general matrix Lie groups, 
by applying \emph{constant-velocity} prior, and defining \emph{locally} linear GP.
This enables using sparse GP approach in a large space of practical SLAM settings.
In this report we give the theory and leave the experimental evaluation in future publications.

\end{abstract}

\section{Introduction}
Simultaneous localization and mapping (SLAM) is a fundamental tool in robotics, 
enabling robots to autonomously operate in previously unseen environments.
Currently, many researchers are focussing on nonlinear optimization techniques
for SLAM. Compared to message-passing techniques like extended Kalman filtering 
and smoothing, nonlinear optimization is able to better contend 
with large scale environments, and sometimes able to produce more consistent results~\cite{Thrun08book}.
For example, batch nonlinear least squares optimization has been widely used
in both the computer vision~\cite{Triggs00} and SLAM communities~\cite{Lu97ar,Thrun06ijrr,Dellaert06ijrr} for years, 
and, recently, incremental approaches to smoothing and mapping~\cite{Kaess08tro,Kaess12ijrr} 
demonstrate that batch solutions can be 
efficiently and incrementally updated as new measurements are collected.

The majority of existing SLAM algorithms use discrete-time representations 
of robot trajectories. Although discrete-time approaches are sufficient for many problems, 
there are two important  situations that discrete-time approaches
have difficulty handling: (1) when sensors measure the environment 
continuously, for example with spinning LIDAR
or rolling-shutter cameras, measurements will be distorted by the robot's motion; and
(2) when sensor measurements arrive asynchrounously.

In these cases, a continuous-time trajectory representation provides a solution. 
Unlike discretized
trajectory representations that are parameterized at fixed discrete time intervals, continuous-time representations can be queried to recover the robot state
at any time of interest. 
Although continuous-time trajectory representations have successfully been used in state estimation~\cite{Kalman61jbe} for years, continuous-time localization and mapping is a recent tool in robotics. 
Several popular continuous-time trajectory representations include 
linear interpolation \cite{Bosse09icra,Li13icra,Dong2014fsr}, 
splines \cite{Bibby10icra,Anderson13icra,Furgale13iros,Leutenegger15ijrr,
Patron15ijcv,Furgale15ijrr}, 
and hierarchical wavelets \cite{Anderson14icra}, all of which have been used in both filtering and batch estimation approaches.

This report focuses on an alternative probabilistic non-parametric representation for trajectories based on Gaussian processes (GPs). 
Tong \textit{et al.}~\cite{Tong13ijrr,Tong14jfr} showed that simultaneous trajectory estimation
and Mapping (STEAM), the continuous-time extension of SLAM, can be reduced to GP regression.
By placing various GP priors on robot trajectories, 
this approach can solve different types of trajectory 
estimation problems. However, if standard kernels, such as the squared exponential kernel, are
used, the method is expensive, with polynomial time and space complexity.

Maintaining sparsity in SLAM problems has been well-studied
\cite{Triggs00,Dellaert06ijrr,Eustice06tro}, and it is the key to maintaining scalable
optimization in many modern SLAM algorithms. 
Barfoot \textit{et al.}~\cite{Barfoot14rss} shows that
by applying a linear time-varying stochastic differential equation 
(LTV-SDE) prior on trajectories, the inverse kernel matrices are exactly sparse, 
leading to efficient GP regression.
This approach is further extended to GP priors driven by
nonlinear time-varying stochastic differential equations (NTV-SDEs) 
\cite{Anderson15ar}, and an incremental GP regression framework~\cite{Yan15isrr}.

The major drawback of all of these Gaussian process-based approaches is they 
require the system state to live in a \emph{vector space}, which is not a
valid assumption in many trajectory estimation problems. For
example, typical vector-valued representations for 3D rigid-body rotations 
either exhibit singularities (Euler angles) or
impose extra nonlinear constraints (quaternions).

Sparse GP regression for STEAM~\cite{Barfoot14rss} has  been extended to 
the special Euclidean group SE(3) in Anderson et al.~\cite{Anderson15iros}.
In this technical report we extend the sparse GP regression approach~\cite{Barfoot14rss} 
to work with arbitrary matrix Lie groups~\cite{Chirikjian11book}, 
extending the approach to a much more general setting.
For example, an attitude and heading reference system
(AHRS)~\cite{Farrell08book} can be treated as trajectory estimation
on the special orthogonal group SO(3), 
and 3D trajectory estimation for a monocular camera 
without scale information can be treated as 
trajectory estimation on the similarity transformation group 
Sim(3)~\cite{Engel14eccv}. 
In this report we only summarize the theory, and leave all the experimental evaluations in future publications.

\section{Preliminaries}

\subsection{Problem Definition}

We consider the problem of 
continuous-time trajectory estimation, in which a continuous-time system state  
$\bm{x}(t)$ is estimated from observations~\cite{Tong13ijrr}.
The system model is described as
\begin{align}
\bm{x}(t) &\sim \mathcal{GP}(\bm{\mu}(t), \bm{\mathcal{K}}(t, t'))  \label{eq:GP_model} \\
\mathbf{z}_i &= \mathbf{h}_i(\bm{x}(t_i))+\mathbf{n}_i, 
\mathbf{n}_i \sim \mathcal{N}(\bm{0}, \mathbf{\Sigma}_i),  \label{eq:meas_model}
\end{align}
where $\bm{x}(t)$ is represented by a Gaussian Process (GP) with mean $\bm{\mu}(t)$
and covariance $\bm{\mathcal{K}}(t, t')$. A measurement $\mathbf{z}_i$ at each time $t_i$ is obtained by the (generally nonlinear) discrete-time measurement function $\mathbf{h}_i$ in Eq.~\eqref{eq:meas_model} and assumed to be corrupted by zero-mean Gaussian noise with covariance $\mathbf{\Sigma}_i$.

\subsection{Maximum a Posteriori Estimation}

The Maximum a Posteriori (MAP) estimate of the trajectory can be computed through  Gaussian process Gauss-Newton (GPGN) \cite{Tong13ijrr}. We first write down the objective function, with assumption that 
there are $M$ observations and the definitions of following terms
\begin{align*}
\bm{x} &\doteq \begin{bmatrix} \bm{x}(t_1) \\ \vdots \\ \bm{x}(t_M) \end{bmatrix}, 
\bm{\mu} \doteq \begin{bmatrix} \bm{\mu}(t_1) \\ \vdots \\ \bm{\mu}(t_M) \end{bmatrix}, 
\bm{\mathcal{K}} \doteq [\bm{\mathcal{K}}(t_i, t_j)]\Bigr|_{ij, 1 \leq i,j \leq M},
\\
\mathbf{z} & \doteq \begin{bmatrix} \mathbf{z}_1 \\ \vdots \\ \mathbf{z}_M \end{bmatrix}, 
\mathbf{h}(\bm{x}) \doteq \begin{bmatrix} \mathbf{h}_1(\bm{x}_1) \\ \vdots \\ 
\mathbf{h}_M(\bm{x}_M) \end{bmatrix}, 
\mathbf{\Sigma} \doteq \begin{bmatrix} \mathbf{\Sigma}_1 & & \\
 & \ddots & \\ & & \mathbf{\Sigma}_M \end{bmatrix},
\end{align*}
the MAP estimation can be written as
\begin{equation}
\bm{x}^{*} = \argmax_{\bm{x}} \bigg\{ \frac{1}{2} \parallel \bm{x} - \bm{\mu}
\parallel^{2}_{\bm{\mathcal{K}}}
+ \frac{1}{2} \parallel \mathbf{h}(\bm{x}) - \mathbf{z} \parallel^{2}_{\bm{\Sigma}}
\bigg\}, \label{eq:MAP}
\end{equation}
where $\parallel \parallel_{\bm{\Sigma}}$ is Mahalanobis distance defined as 
$\parallel \mathbf{x} \parallel^2_{\bm{\Sigma}} \doteq \mathbf{x}^\top
\bm{\Sigma}^{-1} \mathbf{x}$. MAP estimation is therefore translated into
a nonlinear least square optimization problem.

We use a Gauss-Newton approach to solve the nonlinear least squares problem. 
By linearizing the measurement function $\mathbf{h}_i$ around 
a linearization point $\overline{\bm{x}}_i$, we obtain
\begin{equation}
\mathbf{h}_i(\overline{\bm{x}}_i + \delta\bm{x}_i) \approx 
\mathbf{h}_i(\overline{\bm{x}}_i) + \mathbf{H}_i\delta\bm{x}_i, 
\mathbf{H}_i \doteq \frac{\partial\mathbf{h}_i}{\partial\bm{x}}\Bigr|_{\overline{\bm{x}}_i}, 
\end{equation}
in which $\mathbf{H}_i$ is the Jacobian matrix of measurement function \eqref{eq:meas_model}
at linearization point $\overline{\bm{x}}_i$. By defining
$\mathbf{H} \doteq \diag ( \mathbf{H}_1 , \dots , \mathbf{H}_M )$,
we get a linearized least squares problem around linearization point $\overline{\bm{x}}$
\begin{small}
\begin{equation}
\delta\bm{x}^{*} = \argmax_{\delta\bm{x}} \bigg\{ \frac{1}{2} \parallel  
\overline{\bm{x}} + \delta\bm{x} - \bm{\mu} \parallel^{2}_{\bm{\mathcal{K}}}
+ \frac{1}{2} \parallel \mathbf{h}(\overline{\bm{x}}) + \mathbf{H} \delta\bm{x}
- \mathbf{z} \parallel^{2}_{\bm{\Sigma}} \bigg\}. \label{eq:linear_obj_func}
\end{equation}
\end{small}%
%
%
The GPGN algorithm starts from some initial guess of $\overline{\bm{x}}$, 
then, at each iteration, the optimal perturbation $\delta\bm{x}^*$ is found by 
solving linear system 
\begin{equation}
(\bm{\mathcal{K}}^{-1} + \mathbf{H}^\top \mathbf{\Sigma}^{-1} \mathbf{H})\delta\bm{x}^* =
\bm{\mathcal{K}}^{-1}(\bm{\mu} - \overline{\bm{x}}) + 
\mathbf{H}^\top \mathbf{\Sigma}^{-1} (\mathbf{z} - \mathbf{h}). \label{eq:linear_system}
\end{equation}
and updating the solution by 
$\overline{\bm{x}} \leftarrow \overline{\bm{x}} + \delta\bm{x}^*$ 
until convergence.

The information matrix $\bm{\mathcal{K}}^{-1}$ 
in Eq.~\eqref{eq:linear_system} encodes the GP prior information, 
and $\mathbf{H}^\top \mathbf{\Sigma}^{-1} \mathbf{H}$ 
represents information from the measurements. 
 $\mathbf{H}^\top \mathbf{\Sigma}^{-1} \mathbf{H}$ is block-wise sparse
in most SLAM problems~\cite{Dellaert06ijrr}, but $\bm{\mathcal{K}}^{-1}$
is not usually sparse for most commonly used kernels. 
We define a GP prior with sparse structure and exploit this structure to efficiently solve the linear system.

\section{Sparse GP Priors for Trajectory Estimation}

A class of exactly sparse GP priors for trajectory estimation is proposed in Barfoot et al.~\cite{Barfoot14rss}. 
Unfortunately, only vector-valued system states are correctly handled by this approach. 
In this section we first briefly revisit the approach in~\cite{Barfoot14rss} for vector spaces, 
we then extend this approach to two types of Lie groups,
the special orthogonal group SO(3) and the special Euclidean group SE(3).

\subsection{GP Priors for Vector Space} \label{sec:GP_vector}

Here we consider GP priors for vector-valued 
system states $\bm{x}(t)$ generated by linear time-varying 
stochastic differential equations (LTV-SDEs)~\cite{Barfoot14rss}
\begin{equation}
\dot{\bm{x}}(t) = \mathbf{A}(t)\bm{x}(t) + \mathbf{u}(t) + \mathbf{F}(t)\mathbf{w}(t), \label{eq:LTV-SDE}
\end{equation}
where $\mathbf{u}(t)$ is the known system control input, 
$\mathbf{w}(t)$ is white process noise, and both
$\mathbf{A}(t)$ and $\mathbf{F}(t)$ are time-varying system matrices. 
The white process noise is represented by 
\begin{equation}
\mathbf{w}(t) \sim \mathcal{GP}(\mathbf{0}, \mathbf{Q}_C\delta(t-t')), 
\end{equation}
where $\mathbf{Q}_C$ is the power-spectral density matrix, 
which is a hyperparameter~\cite{Anderson15ar},
and $\delta(t-t')$ is the Dirac delta function. 
The mean and covariance of LTV-SDE generated GP are
\vspace{-2mm}
\begin{align}
&\,\,\bm{\mu}(t) = \mathbf{\Phi}(t,t_0) \bm{\mu}_0
+ \int_{t_0}^{t} \mathbf{\Phi}(t,s) \mathbf{u}(s) ds\\
\bm{\mathcal{K}}&(t,t') =  \mathbf{\Phi}(t,t_0) \bm{\mathcal{K}}_0 
\mathbf{\Phi}(t',t_0)^\top\nonumber \\
&+ \int_{t_0}^{\min(t,t')} \mathbf{\Phi}(t,s) \mathbf{F}(s) \mathbf{Q}_C
\mathbf{F}(s)^\top \mathbf{\Phi}(t',s)^\top ds
\end{align}
where $\bm{\mu}_0$ is the initial mean value of first state, 
$\bm{\mathcal{K}}_0$ is the covariance of first state, 
and $\mathbf{\Phi}(t, s)$ is transition matrix.

In \cite{Barfoot14rss} it is proved that if the system 
is generated by the LTV-SDE in Eq.~\eqref{eq:LTV-SDE}, the inverse covariance matrix 
$\bm{\mathcal{K}}^{-1}$ is block-tridiagonal. 

The \emph{constant-velocity} GP prior is generated by a LTV-SDE with white noise on the acceleration and has previously been used 
in trajectory estimation~\cite{Tong13ijrr,Barfoot14rss,Anderson15iros}.
\begin{equation}
\ddot{\mathbf{p}}(t) = \mathbf{w}(t), \label{eq:const-vel-LTV-SDE}
\end{equation} 
where $\mathbf{p}(t)$ is the $N$-dimensional vector-valued position 
(or pose) variable of trajectory, if the system has $N$ degrees of freedom.
To convert this prior into the LTV-SDE form of Eq.~\eqref{eq:LTV-SDE},
a Markov system state variable is declared
\begin{equation}
\bm{x}(t) \doteq \begin{bmatrix} \mathbf{p}(t) \\ \dot{\mathbf{p}}(t) \end{bmatrix},
\end{equation}
The prior in Eq.~\eqref{eq:const-vel-LTV-SDE} then can easily be converted 
into a LTV-SDE in Eq.~\eqref{eq:LTV-SDE} by defining
\begin{equation}
\mathbf{A}(t) = \begin{bmatrix} \mathbf{0} & \mathbf{1} \\ 
\mathbf{0} & \mathbf{0} \end{bmatrix},\quad \mathbf{u}(t) = \mathbf{0},\quad 
\mathbf{F}(t) = \begin{bmatrix} \mathbf{0} \\ \mathbf{1} \end{bmatrix}. 
\end{equation}

\subsection{GP Priors on SO(3)} \label{sec:GP_SO3}

Before discussing sparse GP priors for general Lie groups, 
we discuss several specific examples.
The \emph{Special Orthogonal Group} SO(3), is a matrix Lie group 
and represents 3D rotation matrices $\mathbf{R}$ defined by
$\{\mathbf{R} \in \mathbb{R}^{3 \times 3} : \mathbf{R} \mathbf{R}^\top = \mathbf{I}, 
\det(\mathbf{R}) = 1 \}$.
The continuous-time trajectory is then represented by the function  $\mathbf{R}(t)$ that maps time to rotation matrices.


The relation between rotation and \emph{body-frame angular velocity} is given by 
\cite[p.52]{Murray94book}
\begin{equation}
\dot{\mathbf{R}}(t) = \mathbf{R}(t) _{b}\bm{\omega}(t)^{\wedge}, \label{eq:body-vel-SO3}
\end{equation}
where $_{b}\bm{\omega}(t)$ is the body-frame angular velocity (the subscript $b$ means
the angular velocity is defined in body-frame), 
and $\wedge$ operator constructs a $3 \times 3$ skew symmetric matrix
from a vector in $\mathbb{R}^3$
\begin{equation}
\bm{\omega}^{\wedge} = \begin{bmatrix} \omega_1 \\ \omega_2 \\ \omega_3 \end{bmatrix} 
^{\wedge} = \begin{bmatrix} 0 & -\omega_3 & \omega_2 \\
\omega_3 & 0 & -\omega_1 \\ -\omega_2 & \omega_1 & 0 \end{bmatrix}.
\label{eq:so3-lie-algrebra}
\end{equation} 
Assume that body-frame angular velocity is constant 
(\emph{constant-velocity prior}), and the body-frame angular acceleration is corrupted by white noise 
\begin{equation}
\dot{_{b}\bm{\omega}}(t) = \mathbf{w}(t), \label{eq:SO3_body_noise}
\end{equation}
then we can write down the nonlinear SDE that represents this prior
\vspace{-3mm}
\begin{equation}
\dot{\bm{x}}(t) = \frac{d}{dt} \begin{Bmatrix} \mathbf{R}(t) \\ _{b}\bm{\omega}(t) 
\end{Bmatrix}
= \begin{Bmatrix} \mathbf{R}(t) {}_{b}\bm{\omega}(t)^{\wedge} \\ \mathbf{w}(t) \end{Bmatrix},
\label{eq:SO3_nonlinear_prior}
\end{equation}
where $\bm{x}(t) \doteq \{ \mathbf{R}(t), {}_{b}\bm{\omega}(t) \}$ are 
the Markov system states.
The SDE is nonlinear, so we cannot leverage the approach in \cite{Barfoot14rss}
to get exactly sparse linear system. 
However, similar to approach in~\cite{Anderson15ar,Anderson15iros}, 
it is possible to linearize the system around the current point estimate $\overline{\bm{x}}(t)$, and achieve a \emph{locally} linear SDE, 
which can utilize the exactly sparse GP prior proposed in \cite{Barfoot14rss}.

To define a \emph{locally} linear GP prior, we first look at how to handle uncertainty
and define a locally linear GP on SO(3). 
Various approaches have been proposed to handle uncertainty for SO(3), or even general Lie groups, 
include \cite{Chirikjian11book, Long12rss, Hertzberg2013if, Barfoot2014tro, Forster15rss}. 
Here we adopt the approach in \cite{Forster15rss} and 
define the Gaussian distribution on SO(3) as
a Gaussian distribution on the tangent space that is then mapped back to SO(3) by an 
\emph{exponential map} 
\begin{equation}
\mathbf{\widetilde{R}} = \mathbf{R} \exp(\epsilon^{\wedge}),\quad	
\epsilon \sim \mathcal{N}(\mathbf{0}, \Sigma),  \label{eq:SO3_noise}
\end{equation}
where $\mathbf{\widetilde{R}}$ is noisy rotation, $\mathbf{R}$ is noise-free rotation,
and $\epsilon \in \mathbb{R}^3$ is a small perturbation which is normally distributed 
with zero mean and $\Sigma$ covariance.

By adopting  this definition of a Gaussian distribution on SO(3), 
the GP on SO(3) can be defined locally. Considering a rotation $\mathbf{R}_i$
at time $t_i$ close to $\mathbf{R}(t)$ at $t$ (the time interval 
between $t$ and $t_i$ is small), the GP model in Eq.~\eqref{eq:GP_model} gives
\begin{equation}
\mathbf{R}(t) = \mathbf{R}_i \exp(\bm{\xi}_i(t)^{\wedge}), \quad  \bm{\xi}_i(t) \sim \mathcal{N}(\mathbf{0}, \bm{\mathcal{K}}(t_i, t)).
\label{eq:tan_vec}
\end{equation}
A \emph{local} variable $\bm{\xi}_i(t) \in \mathbb{R}^3$ around $\mathbf{R}_i$ is defined as
\begin{equation}
\bm{\xi}_i(t) \doteq \log(\mathbf{R}_i^{-1} \mathbf{R}(t))^{\vee}, \label{eq:so3_local_var}
\end{equation}
where $\vee$ is the inverse operator of $\wedge$, and $\log(\cdot)$ is the \emph{logarithm map}
of SO(3), which is inverse function of the exponential map. 
The time derivative of $\bm{\xi}_i(t)$ has \cite[p.26]{Chirikjian11book}
\begin{equation}
\mathbf{R}(t)^{-1} \dot{\mathbf{R}}(t) = \big( \bm{\mathcal{J}}_r(\bm{\xi}_i(t))
\dot{\bm{\xi}}_i(t) \big) ^{\wedge},
\end{equation}
where $\bm{\mathcal{J}}_r$ is the \emph{right Jacobian} of SO(3) \cite[p.40]{Chirikjian11book}.
With Eq.~\eqref{eq:body-vel-SO3} we have
\begin{equation}
 \dot{\bm{\xi}}_i(t) = \bm{\mathcal{J}}_r(\bm{\xi}_i(t))^{-1} {}_{b}\bm{\omega}(t).
\end{equation}
If the small time interval assumption is satisfied, since  $\bm{\mathcal{J}}_r$ is 
identity at zero, and $\bm{\xi}_i(t)$ is close to zero, we have a good approximation
of $\dot{\bm{\xi}}_i(t)$
\begin{equation}
\dot{\bm{\xi}}_i(t) \approx  {}_{b}\bm{\omega}(t).
\end{equation}
Here $\dot{\bm{\xi}}_i(t)$ has an explicit meaning: it is the body-frame angular velocity.
Considering the case specified by Eq.~\eqref{eq:SO3_body_noise} where white 
noise is injected into time derivative of $\dot{\bm{\xi}}_i(t)$ by
\begin{equation}
\ddot{\bm{\xi}}(t) = \mathbf{w}(t),
\end{equation} 
the local constant-velocity LTV-SDE of SO(3) is finally written 
\vspace{-1mm}
\begin{equation}
\dot{\bm{\gamma}}_i(t) = \frac{d}{dt} \begin{bmatrix} \bm{\xi}_i(t) \\ \dot{\bm{\xi}}_i(t)
\end{bmatrix}
= \begin{bmatrix} \dot{\bm{\xi}}_i(t) \\ \mathbf{w}(t) \end{bmatrix},
\label{eq:SO3_linear_prior}
\end{equation}
where $\bm{\gamma}_i(t) \doteq [\bm{\xi}_i(t), \dot{\bm{\xi}}_i(t)]^\top$ is 
the local Markov system states around $\mathbf{R}_i$. 
The SDE in Eq.~\eqref{eq:SO3_linear_prior} is linear, so we can apply the approach 
in Section~\ref{sec:GP_vector}. 

\subsection{GP Priors on SE(3)} \label{sec:GP_SE3}

We can define locally linear constant-velocity GP priors for the 
\emph{Special Euclidean Group} SE(3) in similar manner to GP priors on SO(3) in Section~\ref{sec:GP_SO3}.
The Special Euclidean Group SE(3) represents rigid motion in 3D,
which is defined by transformation matrices
\begin{equation}
\mathbf{T} = \begin{bmatrix}
\mathbf{R} & \mathbf{t} \\ \mathbf{0} & 1
\end{bmatrix},
\end{equation}
and where $\mathbf{t}$ is the translation term of motion. 

Similar to SO(3), the body-frame velocity ${}_b\mathbf{v}(t)$ 
has the following relation with the time derivative of transformation matrix
$\dot{\mathbf{T}}(t)$ \cite[p.55]{Murray94book}
\begin{equation}
\dot{\mathbf{T}}(t) = \mathbf{T}(t) {}_b\mathbf{v}(t)^\wedge, \label{eq:body-vel-SE3}
\end{equation}
where $\wedge$ operator constructs a matrix from a velocity $\mathbf{v} \in \mathbb{R}^6$
\begin{equation}
\mathbf{v}^{\wedge} = \begin{bmatrix} \bm{\omega} \\ \bm{v} \end{bmatrix} ^{\wedge} =
\begin{bmatrix} \bm{\omega}^{\wedge} & \bm{v} \\ \mathbf{0} & 0 \end{bmatrix},
\label{eq:se3-lie-algrebra}
\end{equation}
where $\bm{v} = \begin{bmatrix} v_1 & v_2 & v_3 \end{bmatrix}^\top$
is the body-frame translational velocity.
We define the local variable $\bm{\xi}_i(t)$
\begin{equation}
\bm{\xi}_i(t) \doteq \log(\mathbf{T}_i^{-1} \mathbf{T}(t))^{\vee},
\end{equation} 
and similar to SO(3), we get the time derivative of local variable $\bm{\xi}_i(t)$
from Eq.~\eqref{eq:body-vel-SE3} $\dot{\bm{\xi}}_i(t) \approx {}_{b}\mathbf{v}(t)$.
With local variable $\bm{\xi}_i(t)$ and $\bm{\gamma}_i(t)$ defined, 
SE(3) can have locally linear SDE formulated similarly to Eq.~\eqref{eq:SO3_linear_prior},
and an be used to generate a GP prior in the same way.

\section{Sparse GP Priors on Lie Groups}

Discussions in Section \ref{sec:GP_SO3} and \ref{sec:GP_SE3} will be expanded,  
and a \emph{locally} linear GP prior on general real matrix Lie groups 
will be formally defined and discussed in this section.

We begin by providing notation and several definitions.
Every $N$-dimensional matrix Lie group $G$ has an associated \emph{Lie algebra} $\mathfrak{g}$ \cite[p.16]{Chirikjian11book}.
The Lie algebra $\mathfrak{g}$ coincides with the local tangent space 
to the manifold of $G$.
Example Lie algebras of SO(3) and SE(3) are defined
by skew symmetric matrices in Eq.~\eqref{eq:so3-lie-algrebra} 
and Eq.~\eqref{eq:se3-lie-algrebra} respectively.
The \emph{exponential map} $\exp : \mathfrak{g} \rightarrow G$ 
and \emph{logarithm map} $\log : G \rightarrow \mathfrak{g} $ define 
the mapping between the Lie group and Lie algebra respectively \cite[p.18]{Chirikjian11book}.
$G$ also has an associating \emph{hat operator}
$\wedge : \mathbb{R}^N \rightarrow \mathfrak{g}$ and \emph{vee operator}
$\vee :  \mathfrak{g} \rightarrow \mathbb{R}^N$ 
that convert elements in local coordinates $\mathbb{R}^N$ 
to the Lie algebra  $\mathfrak{g}$ and vice versa \cite[p.20]{Chirikjian11book}.

\subsection{Constant-Velocity GP Priors on Lie Groups}

We use $T \in G$ to represent an object in $G$, so the continuous-time 
trajectory is written as $T(t)$, and trajectory states
to be estimated at times $t_1, \dots, t_M$ 
are $T_1, \dots, T_M$.
To perform trajectory estimation on $G$, we first define the Markov 
system states
\begin{equation}
\bm{x}(t) \doteq \{ T(t), \bm{\varpi}(t) \},
\end{equation}
where $\bm{\varpi}(t)$ is the `body-frame velocity' variable defined by
\begin{equation}
\bm{\varpi}(t) \doteq (T(t)^{-1} \dot{T}(t))^{\vee}. \label{eq:lie-body-vel}
\end{equation}
Since $\forall T \in G, \ T^{-1} \dot{T} \in \mathfrak{g}$~\cite[p.20]{Chirikjian11book}, 
we can apply the $\vee$ operator on $T(t)^{-1} \dot{T}(t)$.
In SO(3) and SE(3), $\bm{\varpi}(t)$ is the
body-frame velocity (see Eq.~\eqref{eq:body-vel-SO3}
and \eqref{eq:body-vel-SE3}).
So we call $\bm{\varpi}(t)$ the `body-frame velocity'
in general Lie groups.
The constant `body-frame velocity' prior is defined as
\begin{equation}
\dot{\bm{\varpi}}(t) = \mathbf{w}(t),
\ \ \mathbf{w}(t) \sim \mathcal{GP}(\mathbf{0}, \mathbf{Q}_C\delta(t-t')),
\label{eq:const_vel_gp}
\end{equation}
but this is a nonlinear SDE, which does not match the LTV-SDE defined 
by Eq.~\eqref{eq:LTV-SDE}.

\subsection{Locally Linear Constant-Velocity GP Priors}

To define a LTV-SDE which can leverage the constant-velocity GP prior,
we linearize the Lie group manifold around each $T_i$, 
and define both a \emph{local} GP and LTV-SDE on the linear tangent space.
We first define a local GP for any time $t$ on trajectory 
which meets $t_i \leq t \leq t_{i+1}$, 
\begin{equation}
T(t) = T_i \exp(\bm{\xi}_i(t)^{\wedge}), 
\ \ \bm{\xi}_i(t) \sim \mathcal{N}(\mathbf{0}, \bm{\mathcal{K}}(t_i, t)).
\label{eq:local_const_vel_gp}
\end{equation}
the \emph{local} pose variable $\bm{\xi}_{i}(t) \in \mathbb{R}^N$ 
around $T_i$ is defined by
\begin{equation}
\bm{\xi}_{i}(t) \doteq \log( T_{i}^{-1} T(t) )^{\vee}. \label{eq:local_var}
\end{equation}
The local LTV-SDE that represents constant-velocity information is
\begin{equation}
\ddot{\bm{\xi}_{i}}(t) = \mathbf{w}(t),
\ \ \mathbf{w}(t) \sim \mathcal{GP}(\mathbf{0}, \mathbf{Q}_C\delta(t-t')).
\label{eq:const_vel_sde}
\end{equation}
If we define the local Markov system state
\begin{equation}
\bm{\gamma}_i(t) \doteq \begin{bmatrix}
\bm{\xi}_{i}(i) \\ \dot{\bm{\xi}_{i}}(t)
\end{bmatrix},
\label{eq:markov_x}
\end{equation}
the local LTV-SDE is rewritten as
\begin{equation}
\dot{\bm{\gamma}}_i(t) = \frac{d}{dt} \begin{bmatrix} \bm{\xi}_i(t) \\ \dot{\bm{\xi}}_i(t)
\end{bmatrix}
= \begin{bmatrix} \dot{\bm{\xi}}_i(t) \\ \mathbf{w}(t) \end{bmatrix}.
\label{eq:const_vel_sde_x}
\end{equation}
To prove the equivalence between the nonlinear SDE in Eq.~\eqref{eq:const_vel_gp}
and the local LTV-SDE in Eq.~\eqref{eq:const_vel_sde}, 
we first look at \cite[p.26]{Chirikjian11book}
\begin{equation}
T(t)^{-1} \dot{T}(t) = \big( \bm{\mathcal{J}}_r(\bm{\xi}_i(t))
\dot{\bm{\xi}}_i(t) \big) ^{\wedge},
\end{equation}
where $\bm{\mathcal{J}}_r$ is the \emph{right Jacobian} of $G$.
With Eq.~\eqref{eq:lie-body-vel} we have
\begin{equation}
 \dot{\bm{\xi}}_i(t) = \bm{\mathcal{J}}_r(\bm{\xi}_i(t))^{-1} \bm{\varpi}(t).
\end{equation}
If the small time interval assumption between any $t_i$ and $t_{i+1}$ 
is satisfied, we have a good approximation
of $\dot{\bm{\xi}}_i(t)$
\begin{equation}
\dot{\bm{\xi}}_i(t) \approx  \bm{\varpi}(t).
\end{equation}
So we have proved that the LTV-SDE in Eq.~\eqref{eq:const_vel_sde}
and \eqref{eq:const_vel_sde_x} is a 
good approximation of constant `body-frame velocity' prior 
defined by Eq.~\eqref{eq:const_vel_gp}.Note that Section~\ref{sec:GP_SO3} 
is a specialization of the above discussion to SO(3).

Both the local GP and LTV-SDE are defined on the tangent space, 
so they are only valid around current linearization points $T_i$.
But if all time stamps have a small enough interval,
the GP and LTV-SDE can be defined in a piecewise manner, 
and every point on the trajectory can be converted to local variable $\bm{\xi}_i(t)$
based on its nearby estimated state $T_i$. 

\subsection{A Factor Graph Perspective}

Once the local GP and constant-velocity LTV-SDE are defined, 
we can write down the cost function $J_{gp}$ used to incorporate information about the GP prior into  
the nonlinear least squares optimization in Eq.~\eqref{eq:MAP}.
As discussed, the GP prior cost function has the generic form
\begin{equation}
J_{gp} = \frac{1}{2} \parallel \bm{\mu} - \bm{x} \parallel^{2}_{\bm{\mathcal{K}}},
\end{equation}
but if the trajectory is generated by a constant-velocity LTV-SDE 
in Eq.~\eqref{eq:const_vel_sde_x}, the GP prior cost can be specified as 
as~\cite{Barfoot14rss}
\begin{align}
J_{gp} & = \sum_{i} \frac{1}{2} \mathbf{e}_i^\top \mathbf{Q}_i^{-1} \mathbf{e}_i, \\
\mathbf{e}_i &= \mathbf{\Phi}(t_{i+1}, t_i) \bm{\gamma}_i(t_i) - \bm{\gamma}_i(t_{i+1}),
\end{align}
where $\mathbf{Q}_i$ is covariance matrix by~\cite{Barfoot14rss}
\begin{equation}
\mathbf{\Phi}(t, s) = \begin{bmatrix} \mathbf{1} & (t-s) \mathbf{1} \\
\mathbf{0} & \mathbf{1} \end{bmatrix}, 
\mathbf{Q}_i = \begin{bmatrix} \frac{1}{3} \Delta t_i^3 \mathbf{Q}_C &
\frac{1}{2} \Delta t_i^2 \mathbf{Q}_C \\ 
\frac{1}{2} \Delta t_i^2 \mathbf{Q}_C &
\Delta t_i \mathbf{Q}_C \end{bmatrix},
\end{equation}
where $\Delta t_i = t_{i+1} - t_i$.

\begin{figure}
\begin{centering}
{\includegraphics[width=0.75\columnwidth]{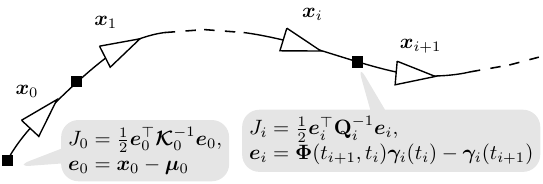}}
\par\end{centering}
\protect\caption{
An example factor graph,   
showing states (triangles) and factors (black boxes).
GP prior factors connect consecutive states,
and define the prior information on first state.
\vspace{-8mm}
\label{fig:factor_graph}}
\end{figure}

Since the GP prior cost $J_{gp}$ has been written as a sum of squared cost terms, 
and each cost term is only related to nearby (local) Markov states,
we can represent the least square problem by \emph{factor graph} models.
In factor graphs the system states are represented by variable factors, 
and the cost terms are represented by cost factors.
An example factor graph is shown in Fig.~\ref{fig:factor_graph}. 
By converting nonlinear least squares problems into factor graphs
we can take advantage of factor graph inference tools 
to solve the problems efficiently.
Additional information about the relationship between factor graphs 
and sparse GP and SLAM problems can be found in 
\cite{Dellaert06ijrr, Barfoot14rss, Anderson15ar, Yan15isrr}

\subsection{Querying the Trajectory}

One of the advantages of representing the continuous-time trajectory as  a GP is that 
we have the ability to query the state of the robot at any time along the trajectory.
For constant-velocity GP priors, the system state 
$\bm{x}(\tau), t_i \leq \tau \leq t_{i+1}$ can be estimated
by two nearby states $\bm{x}(t_i)$ and $\bm{x}(t_{i+1})$~\cite{Barfoot14rss}, 
which allows efficient $O(1)$ interpolation.
We first calculate the mean value of local state $\hat{\bm{\gamma}}_i(\tau)$
\begin{equation}
\hat{\bm{\gamma}}_{i}(\tau) = 
\mathbf{\Lambda}(\tau) \hat{\bm{\gamma}}_{i}(t_i) + 
\mathbf{\Psi}(\tau) \hat{\bm{\gamma}}_{i}(t_{i+1}), \label{eq:interpolate_local} \vspace{-2mm}
\end{equation}
\vspace{-2mm}%
where
\begin{align}
\mathbf{\Lambda}(\tau) &= \mathbf{\Phi}(\tau, t_{i}) - 
\mathbf{Q}_{\tau} \mathbf{\Phi}(\tau, t_{i})^\top \mathbf{Q}_{i+1}^{-1}
\mathbf{\Phi}(t_{i+1}, t_{i}),
\\
\mathbf{\Psi}(\tau) &= \mathbf{Q}_{\tau} \mathbf{\Phi}(\tau, t_{i})^\top
\mathbf{Q}_{i+1}^{-1}.
\end{align}
Once we have the mean value of local state $\hat{\bm{\gamma}}_{i}(\tau)$, the mean value of 
the full state $\hat{\bm{x}}(\tau) = \{ \hat{T}(\tau), \hat{\bm{\varpi}}(\tau) \}$ is
\begin{small}
\begin{align}
\hat{T}(\tau) &= \hat{T}_i \exp \Big( \big( 
\mathbf{\Lambda}_1(\tau) \hat{\bm{\gamma}}_{i}(t_i) + 
\mathbf{\Psi}_1(\tau) \hat{\bm{\gamma}}_{i}(t_{i+1}) \big)^{\wedge} \Big), 
\label{eq:interpolate_pose}
\\
\hat{\bm{\varpi}}(\tau) &= 
\bm{\mathcal{J}}_r \big( \hat{\bm{\xi}}_{i}(\tau) \big) ^{-1} 
\big(
\mathbf{\Lambda}_2(\tau) \hat{\bm{\gamma}}_{i}(t_i) + 
\mathbf{\Psi}_2(\tau) \hat{\bm{\gamma}}_{i}(t_{i+1})
\big),
\label{eq:interpolate_vel}
\end{align}
\end{small}%
where
\vspace{-1mm}
\begin{align*}
\mathbf{\Lambda}(\tau) &= \begin{bmatrix}
\mathbf{\Lambda}_1(\tau) \\ \mathbf{\Lambda}_2(\tau)
\end{bmatrix}, \,\,
\mathbf{\Psi}(\tau) = \begin{bmatrix}
\mathbf{\Psi}_1(\tau) \\ \mathbf{\Psi}_2(\tau)
\end{bmatrix}, 
\\
\hat{\bm{\gamma}}_{i}(t_i) &= \begin{bmatrix}
\mathbf{0} \\ \hat{\bm{\varpi}}(t_i)
\end{bmatrix},\,\,
\hat{\bm{\gamma}}_{i}(t_{i+1}) = \begin{bmatrix}
\hat{\bm{\xi}}_{i}(t_{i+1}) \\ 
\bm{\mathcal{J}}_r(\hat{\bm{\xi}}_{i}(t_{i+1}))^{-1} \hat{\bm{\varpi}}(t_{i+1})
\end{bmatrix},
\\
\hat{\bm{\xi}}_{i}(\tau) &= \log( \hat{T}_i^{-1} \hat{T}(\tau) ) ^ \vee,\quad
\hat{\bm{\xi}}_{i}(t_{i+1}) = \log( \hat{T}_i^{-1} \hat{T}_{i+1} ) ^ \vee,
\end{align*}
%

\subsection{Fusion of Asynchronous Measurements}


Continuous-time trajectory interpolation affords GP-based trajectory estimation methods
several advantages over discrete-time localization algorithms.  In addition to
providing a method for querying the trajectory at any time of interest, GP interpolation
can be used to reduce the number of states needed to represent the robot's trajectory, and 
elegantly handle asynchronous measurements.

\begin{figure}
\centering
\begin{subfigure}[b]{0.22\textwidth}
\centering
\includegraphics[width=1\linewidth]{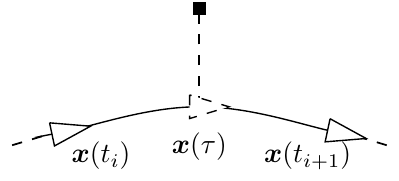}
\vspace*{-6mm}
\caption{Meausurement}
\end{subfigure}
\quad
\begin{subfigure}[b]{0.22\textwidth}
\centering
\includegraphics[width=1\linewidth]{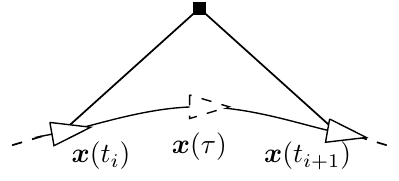}
\vspace*{-6mm}
\caption{Interpolated Factor}
\end{subfigure}
\protect\caption{
(a) Measurement at time $\tau$, 
dashed line indicates it's not an actual factor. 
(b) The interpolated factor encodes measurement at time $\tau$.
\vspace{-3mm}
\label{fig:interpolated_factor}}
\end{figure}

Assume there is a measurement $\mathbf{z}_{\tau}$ of state $\bm{x}(\tau)$ available at an arbitrary time 
$\tau, t_i \leq \tau \leq t_{i+1}$,
with measurement function $\mathbf{h}_{\tau}$ 
and corresponding covariance $\mathbf{\Sigma}_{\tau}$.
The measurement cost in the factor graph can be written as
\begin{equation}
J_{\tau}(\bm{x}(\tau)) = \frac{1}{2} \parallel \mathbf{z}_{\tau} 
- \mathbf{h}_{\tau}(\bm{x}(\tau)) \parallel^{2}_{\mathbf{\Sigma}_{\tau}}.
\end{equation}
Since system state $\bm{x}(\tau)$ is not explicitly available during optimization,
we perform trajectory interpolation between $\bm{x}_{i}$ and $\bm{x}_{i+1}$
by Eq.~\eqref{eq:interpolate_pose} -- \eqref{eq:interpolate_vel},
and rewrite the cost in terms of the interpolated mean value $\hat{\bm{x}}(\tau)$
\begin{equation}
J_{\tau}(\bm{x}_{i}, \bm{x}_{i+1}) = \frac{1}{2} \parallel \mathbf{z}_{\tau} 
- \mathbf{h}_{\tau}(\hat{\bm{x}}(\tau)) \parallel^{2}_{\mathbf{\Sigma}_{\tau}}.
\end{equation}
Because the measurement cost is represented by $\bm{x}_{i}$ and $\bm{x}_{i+1}$,
a binary factor can be added to the factor graph and optimized without explicitly adding an additional state. 
An example is illustrated in Fig.~\ref{fig:interpolated_factor}.

\subsection{Simultaneous Trajectory Estimation and Mapping}
\begin{figure}
\begin{centering}
{\includegraphics[width=0.75\columnwidth]{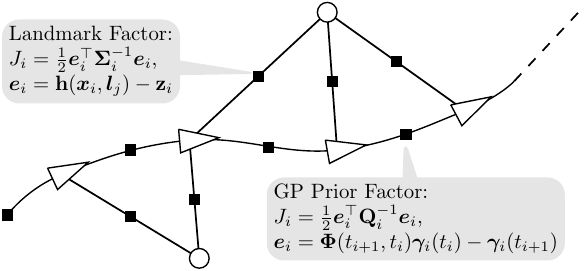}}
\par\end{centering}
\protect\caption{
A factor graph of an example STEAM problem containing GP prior factors and landmark measurements factors.
Landmarks are illustrated with open circles.
\label{fig:factor_graph_steam}}
\end{figure}

The proposed approach could be extended 
from trajectory estimation
to Simultaneous Trajectory Estimation and Mapping (STEAM) by combining landmarks $\bm{l}$ 
with trajectory states $\bm{x}$, resulting a combined state 
$\bm{z} \doteq [\bm{x}, \bm{l}]^\top, 
\bm{l} \doteq [\bm{l}_1, \dots, \bm{l}_L]^\top$,
where $L$ is number of landmarks~\cite{Barfoot14rss}. The resulting linear system has the form
\begin{equation}
\underbrace{\begin{bmatrix}
\mathbf{W}_{xx} & \mathbf{W}_{lx}^{\top} \\ \mathbf{W}_{lx} & \mathbf{W}_{ll} 
\end{bmatrix}}_\text{$\mathbf{W}$} 
\underbrace{\begin{bmatrix}
\delta\bm{x}^{*} \\ \delta\bm{l}^{*} 
\end{bmatrix}}_\text{$\delta\bm{z}^{*} $} 
= \underbrace{\begin{bmatrix}
\bm{b}_x \\ \bm{b}_l 
\end{bmatrix}}_\text{$\bm{b}$},
\end{equation}
where $\mathbf{W}_{xx}$ is block-tridiagonal due to the GP prior,
$\mathbf{W}_{ll}$ is block-diagonal, and $\mathbf{W}_{lx}$ depends
on landmark observations, but is generally sparse \cite{Dellaert06ijrr}.
Block-wise sparse Cholesky decomposition can be applied to solve 
the linear system
\begin{equation}
\underbrace{\begin{bmatrix}
\mathbf{V}_{xx} & \mathbf{0} \\ \mathbf{V}_{lx} & \mathbf{V}_{ll} 
\end{bmatrix}}_\text{$\mathbf{V}$} 
\underbrace{\begin{bmatrix}
\mathbf{V}_{xx}^\top & \mathbf{V}_{lx}^\top \\ \mathbf{0}& \mathbf{V}_{ll}^\top 
\end{bmatrix}}_\text{$\mathbf{V}^\top$} 
= \underbrace{\begin{bmatrix}
\mathbf{W}_{xx} & \mathbf{W}_{lx}^{\top} \\ \mathbf{W}_{lx} & \mathbf{W}_{ll} 
\end{bmatrix}}_\text{$\mathbf{W}$}.
\end{equation}
%
A factor graph representing an example STEAM problem is shown in Fig.~\ref{fig:factor_graph_steam}.

\section{Conclusion}

We extend the sparse GP regression approach of Barfoot et al.~\cite{Barfoot14rss} to 
general matrix Lie groups.
As a result, we can solve a broader class of trajectory estimation problems,
such as attitude and trajectory estimation in 3D space.
Additionally, the continuous-time trajectory representation gives us the 
ability to fuse asynchronous sensor measurements during estimation, which is useful in practice.

Although the proposed method is defined and evaluated in batch settings,
it can be immediately extended to an incremental estimation framework
by combining the proposed sparse GP priors with the incremental GP regression
framework of Yan et al.~\cite{Yan15isrr}.
In fact, the proposed GP prior is not limited 
to just STEAM problems: any technique that benefits from continuous-time trajectories represented
by sparse GPs can use our priors. For example, 
by using our approach in motion planning \cite{Mukadam16icra,Dong16rss}
one could immediately extend vector-space motion planning algorithms to
motion planning algorithms on general matrix Lie groups. 

\bibliographystyle{ieeetr}
\bibliography{refs}

\end{document}